\documentclass[10pt,twocolumn,letterpaper]{article}

\usepackage[pagenumbers]{cvpr} 

\usepackage[dvipsnames]{xcolor}

\newcommand{\Supp}{supp. mat.~}

\definecolor{cvprblue}{rgb}{0.21,0.49,0.74}
\usepackage[pagebackref,breaklinks,colorlinks,citecolor=cvprblue]{hyperref}
\usepackage{algorithm}
\usepackage{algpseudocode}
\usepackage{amsmath}
\usepackage{amssymb}
\usepackage{booktabs}
\usepackage{mathtools}
\usepackage{multirow}

\usepackage{graphicx} 
\usepackage{adjustbox} 

\DeclareMathOperator*{\argmin}{arg\,min}

\def\paperID{4587} 
\def\confName{CVPR}
\def\confYear{2024}

\title{Optimizing Diffusion Noise Can Serve As Universal Motion Priors}

\newcommand*{\affmark}[1][*]{\textsuperscript{#1}}

\author{
Korrawe Karunratanakul\affmark[1] \quad
Konpat Preechakul\affmark[2] \quad
Emre Aksan\affmark[4] \\ 
Thabo Beeler\affmark[4] \quad
Supasorn Suwajanakorn\affmark[3] \quad
Siyu Tang\affmark[1] 
\vspace{0.7em} \\
\affmark[1]{ETH Z{\"u}rich, Switzerland} \quad
\affmark[2]{UC Berkeley} \\
\affmark[3]{VISTEC, Thailand}  \quad
\affmark[4]{Google}  \\
{\small\url{https://korrawe.github.io/dno-project/}}  \vspace{-0.5em}
}

\begin{document}
\newcommand{\myparagraph}[1]{\vspace{0.5em}\noindent\textbf{#1}}
\newcommand{\ST}[1]{{\color{red}[ST: #1]}}
\newcommand{\KK}[1]{{\color{magenta}[KK: #1]}} 
\newcommand{\EA}[1]{{\color{orange}[EA: #1]}}
\newcommand{\KP}[1]{{\color{blue}[KP: #1]}}
\newcommand{\kp}[1]{\KP{#1}}
\newcommand{\st}[1]{\ST{#1}}
\newcommand{\ea}[1]{\EA{#1}}
\newcommand{\V}[1]{\mathbf{#1}}
\newcommand{\R}[0]{\rm I\!R}
\newcommand{\E}[0]{\rm I\!E}
\newcommand{\loss}[0]{\mathcal{L}}

\newcommand{\methodname}{{DNO}}
\newcommand{\fullname}{{Diffusion Noise Optimization}}

\newcommand{\vaename}{{???~}}

\newcommand{\norm}[1]{\left\lVert#1\right\rVert}

\newcommand{\cmark}{\ding{51}}%
\newcommand{\xmark}{\ding{55}}%

\newcommand{\xt}{{\mathbf{x}_t}}
\newcommand{\xT}{{\mathbf{x}_T}}
\newcommand{\xtone}{{\mathbf{x}_{t-1}}}
\newcommand{\xzero}{{\mathbf{x}_0}}
\newcommand{\x}{\mathbf{x}}
\newcommand{\y}{\mathbf{y}}
\newcommand{\m}{\mathbf{m}}
\newcommand{\z}{\mathbf{z}}
\newcommand{\I}{\mathbf{I}}
\newcommand{\defeq}{\vcentcolon=}

\twocolumn[{%
\renewcommand\twocolumn[1][]{#1}%
\maketitle

\begin{center}
  \newcommand{\teaserwidth}{\textwidth}
  \vspace{-0.3cm}
  \centerline{\includegraphics[width=\linewidth]{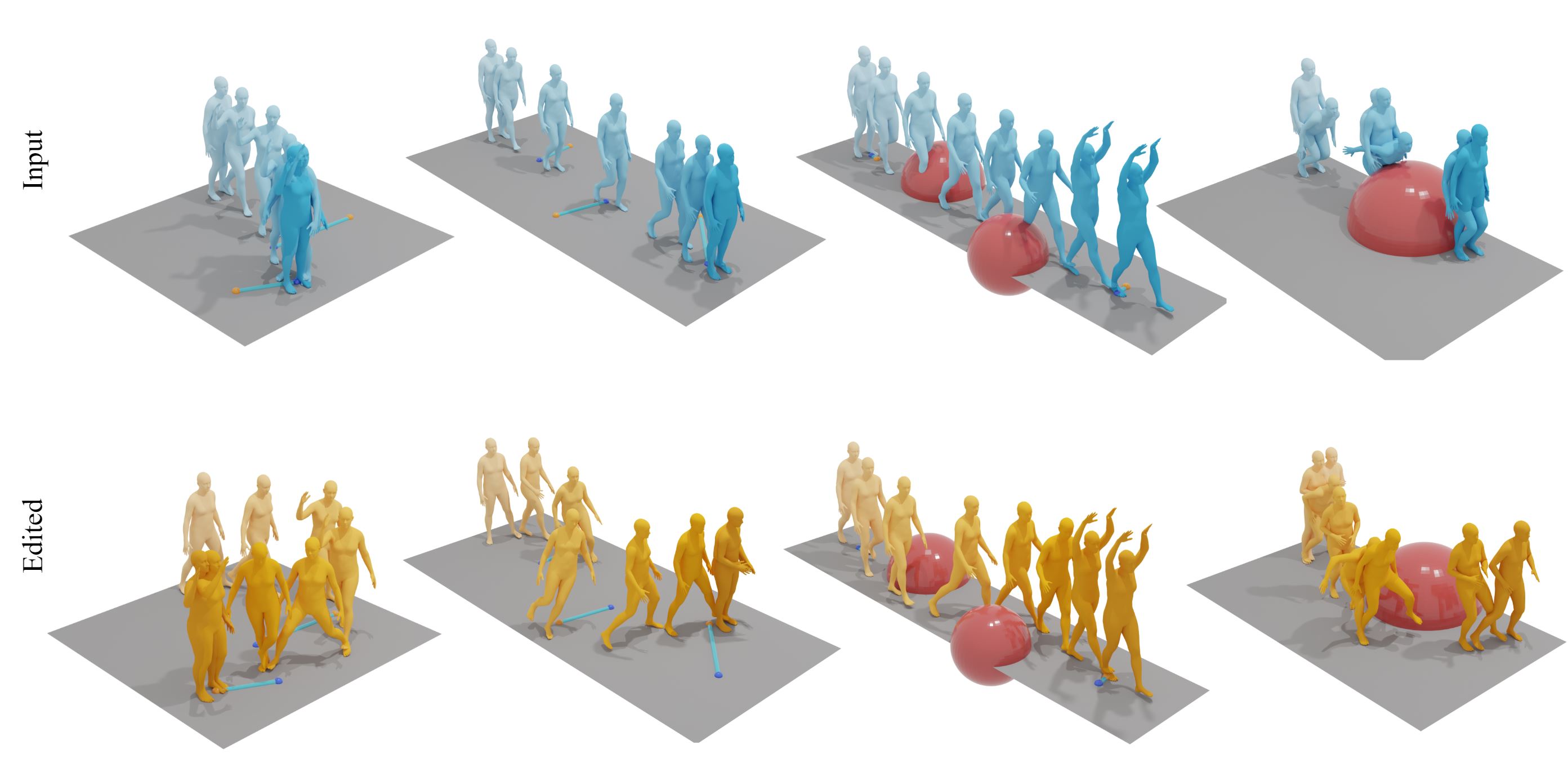}}
    \captionof{figure}{Our proposed \fullname\ (\methodname) can leverage the existing human motion diffusion models as universal motion priors. We demonstrate its capability in the motion editing tasks where \methodname~can preserve the content of the original model and accommodates a diverse range of editing modes, including changing trajectory, pose, joint location, and avoiding newly added obstacles.
    }
    \label{gi:teaser}
\end{center}%
}]

\begin{abstract}
We propose \fullname\ (\methodname),
a new method that effectively leverages existing motion diffusion models as motion priors for a wide range of motion-related tasks.
Instead of training a task-specific diffusion model for each new task, \methodname~operates by optimizing the diffusion latent noise of an existing pre-trained text-to-motion model.
Given the corresponding latent noise of a human motion, it propagates the gradient from the target criteria defined on the motion space through the whole denoising process to update the diffusion latent noise.
As a result, \methodname~supports any use cases where criteria can be defined as a function of motion. 
In particular, we show that, for motion editing and control, \methodname~outperforms existing methods in both achieving the objective and preserving the motion content.
\methodname~accommodates a diverse range of editing modes, 
including changing trajectory, pose, joint locations, or avoiding newly added obstacles.
In addition, \methodname~is effective in motion denoising and completion, producing smooth and realistic motion from noisy and partial inputs.
\methodname~achieves these results at inference time without the need for model retraining, offering great versatility for any defined reward or loss function on the motion representation.

\end{abstract}    
\section{Introduction}

Many applications of great interest to the motion modeling community can be framed as ``finding a plausible motion that fulfills a set of criteria'', typically formulated as minimizing a cost function addressing the given constraints.
These include but are not limited to generating motions that follow a trajectory, target locations for joints, keyframes, or avoiding obstacles; denoising, completing missing parts of a motion or editing an existing one. Though they may seem diverse, a unified framework should be able to address these diverse yet relevant tasks, which typically follow a task-agnostic motion prior that ensures plausible motions and task-specific cost functions. Such high-quality motion priors are highly sought after, often with efforts that focus on improving architectures. Yet, a truly versatile framework utilizing the underlying motion prior expressively is still lacking.

Among many human motion priors, the diffusion-based models \cite{Li2023-dn,Shafir2024-qh, ma2022mofusion, Karunratanakul2023-nb,Xie2024-zz,Du2023-kh} have become the most prominent ones by achieving impressive performance on certain motion modeling tasks. 
Avatar Grow Legs \cite{Du2023-kh} and EgoEgo \cite{Li2023-dn} can generate full-body motions from limited joint specifications, such as head and hand poses, but their specific scope precludes their use as general motion priors.
PriorMDM \cite{Shafir2024-qh} introduces a more flexible setting; however, the underlying root-relative motion representation requires dense condition signals for individual joints or keyframes, limiting its use to a small set of applications.
OmniControl \cite{Xie2024-zz} addresses this limitation by augmenting the motion diffusion model with another network to encode input conditions. While this improves the fidelity and quantity of control signals, it is not always straightforward to support many other constraints.
For instance, generating locomotion in a scene with obstacles requires a perception of the physical scene, which is unclear how to cover arbitrary scenes effectively. It is only exacerbated when the scene dynamically changes over time, such as a scene populated by other human agents.
GMD \cite{Karunratanakul2023-nb} presents a more versatile approach by limiting the conditioning to task-based objectives, which theoretically supports arbitrary use cases. Yet there is a trade-off between this flexibility and the fidelity of controls and quality of motions.

\begin{figure}[h]
    \centering
    \begin{subfigure}[b]{\linewidth}
        \includegraphics[width=\textwidth]{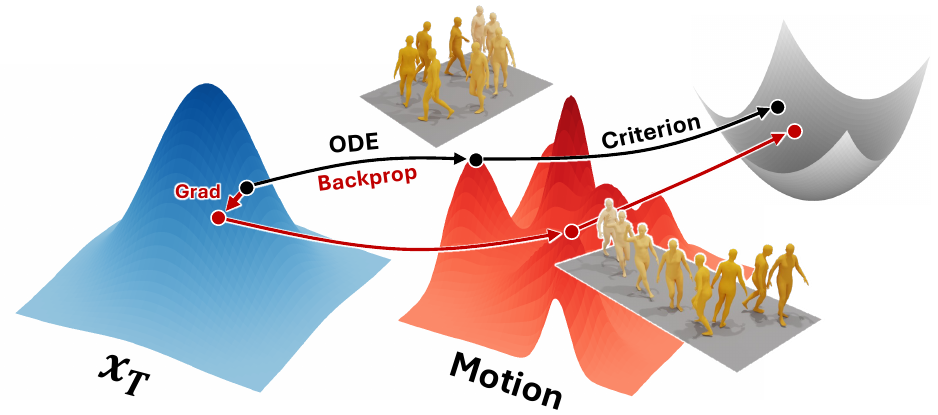}
        \caption{At each optimization step, \methodname~maintains the output motion equality by making a step in the latent space $\xT$, which is decodable to a realistic motion almost everywhere. }
    \label{fig:pipeline}
    \end{subfigure}
    
    \vspace{1em} 
        \begin{subfigure}[b]{\linewidth}
        \includegraphics[width=\textwidth]{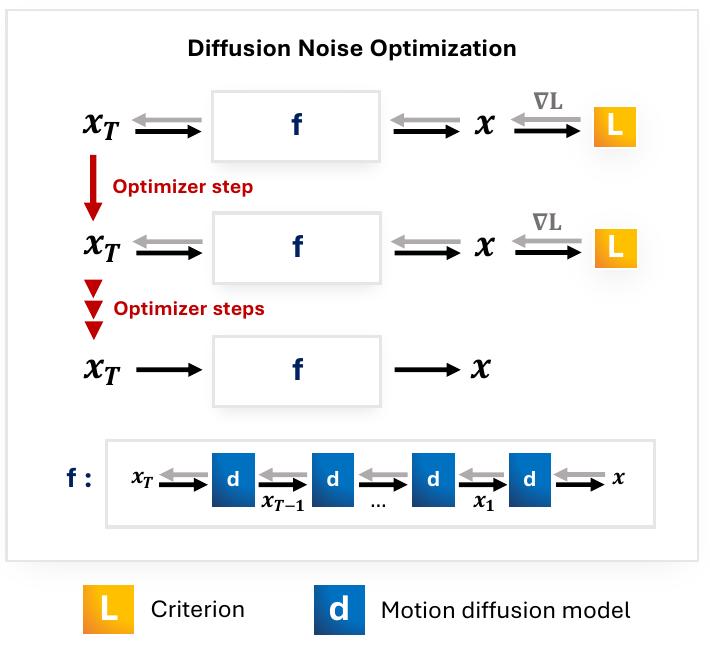}
        \caption{\methodname's step direction is obtained from the gradient of a task-specific criterion $\mathcal{L}$ via backpropagation through an ODE solver, $f(\xT) = \mathrm{ODESolver}(d(\cdot), \xT)$. 
        At convergence, the optimized $x_T$ is ultimately decoded via the ODE solver to get the prediction $x_0$.}
        \label{fig:blocks}
    \end{subfigure}
    \caption{\fullname~(\methodname).}
    \label{fig:method}
    \vspace{-0.5em}
\end{figure}

To this end, we propose a simple yet effective approach to utilize a motion diffusion model as a motion prior. By treating the denoising process as a black box, we can frame motion-related tasks, such as motion editing and refinement, as optimization problems on the latent manifold of the diffusion model, similar to other classes of generative models such as GANs \cite{pan2023drag} and VAEs \cite{Rempe2021-hp} where optimization is seen as iteratively updating the solution on the learned motion manifold (Fig.~\ref{fig:method}a).
In this work, we demonstrate that it is possible and feasible to back-propagate gradients through the full-chain diffusion process, and then optimize the noise vector based on user-provided criteria in the motion space.
This simple approach, dubbed \textbf{\fullname\ (\methodname)}, which we show to be effective at editing and preserving contents, is a unified and versatile framework that supports a wide variety of applications without the need for fine-tuning the underlying motion model for each specific application.
By changing the optimization objective, defined as any differentiable loss function computed on the output motion, \methodname~enables the diffusion model to effectively serve as a motion prior.

Our experiments show that our unified framework \methodname, without any model fine-tuning, produces high-quality motions for motion editing, outperforming existing methods in both preserving the motion content and fulfilling a wide range of objectives including changing trajectory, pose, joint location, and avoiding obstacles. In addition, we demonstrate that it can produce smooth and realistic motion from noisy and partial inputs.
Lastly, we provide extensive studies to validate our design choices, which can serve as a basis to effectively extend our framework to other motion-related tasks.

\section{Related Works}

\myparagraph{Motion synthesis, editing, and completion.}
The human motion generation task aims to generate motions either conditionally or unconditionally \cite{safonova07motiongraphs,pavlakos2019expressive,yan2019convolutional,zhang2020perpetual,zhao2020bayesian}.
Various conditioning signals have been explored such as partial poses \cite{duan2021single,harvey2020robust}, trajectories \cite{wang2021synthesizing,zhang2021learning,kaufmann2020convolutional, Shafir2024-qh}, images \cite{Rempe2021-hp,chen2022learning}, music \cite{lee2019dancing,li2022danceformer,li2021ai}, objects \cite{wu2022saga}, action labels \cite{petrovich2021action,guo2020action2motion}, scene \cite{huang2023diffusion}, or text \cite{ahuja2019language2pose,guo2022t2m,guo2022tm2t,kim2022flame,petrovich2022temos}. 
Recently, the focus has been on generating motion based on natural-language descriptions using diffusion-based models \cite{Tevet2023-ih, Chen2023-id}.
These models, utilizing the CLIP model \cite{radford2021learning}, have shown significant improvements in text-to-motion generation \cite{chen2022mld,yuan2022physdiff,zhang2022motiondiffuse} and support conditioning on partial motions or music \cite{tseng2022edge,alexanderson2022listen}.
Alternatively, the motion can be treated as a new language and embedded into the language model framework \cite{zhang2023t2mgpt,jiang2023motiongpt}
Nevertheless, they lack the ability to handle spatial conditioning signals, such as keyframe locations or trajectories. 
GMD \cite{Karunratanakul2023-nb} handles this problem with classifier-based guidance at inference time to steer the denoising process toward the target conditions. OmniControl \cite{Xie2024-zz} combined GMD with ControlNet \cite{zhang2023controlnet} to improve realism but limits the conditioning signals to text and partial motion observations.
However, motion editing under the motion synthesis framework remains unexplored due to the lack of an explicit mechanism to retain the original motion content.

In parallel, the generation of full-body poses from sparse tracking signals of body joints has gained considerable interest within the community.
Previous work such as EgoEgo \cite{Li2023-dn}, AGRoL \cite{Du2023-kh}, and AvatarPoser \cite{jiang2022avatarposer} while showing impressive results, tend to be specialized models trained explicitly for the motion completion or denoising task.
Therefore, it is unclear how we can leverage the motion priors in these trained models for solving other tasks.
In this work, we focus on using the trained motion diffusion model to tackle various motion-related tasks under a unified framework including editing, completion, and refinement.

\myparagraph{Diffusion models and guidance.}
Diffusion-based probabilistic generative models (DPMs), a class of generative models based on learning to progressively denoising the input data, \cite{Ho2020-ew, Sohl-Dickstein2015-hp, Song2019-xr, Song2021-yj} have gained significant attention across multiple fields of research. They have been successfully applied to tasks such as image generation \cite{Dhariwal2021-bt}, image super-resolution \cite{Saharia2021-lm, Li2022-qr}, speech synthesis \cite{Kong2021-sn, Popov2021-qe}, video generation \cite{Ho2022-qm, Ho2022-gh}, 3D shape generation \cite{Poole2022-kx, Watson2022-vm}, and reinforcement learning \cite{Janner2022-no}.
The growing interest in the diffusion-based model stems from their superior results and impressive generation controllability, for example, in text-conditioned generation \cite{Ramesh2022-id,Rombach2022-nu,Saharia2022-ns} and image-conditioned editing \cite{Meng2022-bn,Choi2021-oe,Brooks2023-wx,Hertz2022-td,Balaji2022-gh}.
%
In terms of conditioning, there are various methods for the diffusion-based models such as imputation and inpainting \cite{Chung2022-tm,Choi2021-oe,Meng2022-bn}, classifier-based guidance \cite{Dhariwal2021-bt, Chung2022-tm}, and classifier-free guidance \cite{Rombach2022-nu,Saharia2022-ns,Ramesh2022-id, ho2021classifierfree}. 
For refinement, SDEdit \cite{Meng2022-bn} enables the repetition of the denoising process to gradually improve output quality but lacks the ability to provide editing guidance.
%
Recently, DOODL \cite{Wallace2023-ng} demonstrates a direct latent optimization approach for image editing with the help of an invertible ODE \cite{Wallace2023-ja}.
Being inspired by that work, we propose a related method with an improved optimization algorithm that speeds up optimization. \methodname~can effectively be used in various motion-related tasks making it suitable as a versatile human motion priors. Additionally, we discover that an invertible ODE is not required which makes \methodname~simpler and convenient to use with minimal effort.

\section{Background}

\subsection{Motion generation with diffusion model}
A diffusion probabilistic model is a denoising model that learns to invert a diffusion process.
A diffusion process is defined as $q(\xt|\xzero) = \mathcal{N}(\sqrt{\alpha_t}\xzero, (1-\alpha_t)\mathbf{I})$ where $\xzero$ is a clean motion and $\xt$ is a noisy motion at the level of $t$ defined by noise schedule $\alpha_t$.
With the diffusion process, we can infer an inverse denoising process $q(\xtone|\xt, \xzero)$. 
Now, we can train the diffusion model by predicting $q(\xtone|\xt, \xzero)$ with a learned $p_\theta(\xtone|\xt, c)$, which is parameterized by a function $d_\theta(\xt, c)$ where $c$ is additional conditions, e.g.~text prompts.
Conditioning is particularly useful for specifying what kind of motion activity we want from the model. 

While diffusion models are stochastic, 
there exist deterministic sampling processes that share the same marginal distribution. These processes include those defined by probability flow ODE \cite{Song2021-yj} or by reformulating the diffusion process to be non-Markovian as in DDIM \cite{Song2021-ac}.
To obtain a sample based on deterministic sampling, we can solve the associated ODE using an ODE solver starting from $\xT \sim \mathcal{N}(\mathbf{0}, \mathbf{I})$.

\subsection{Diffusion model inversion} \label{sec:ddim_inversion}
Diffusion inversion is a process that retrieves the corresponding noise map $\xT$ given an input $\xzero$.
Under deterministic sampling, the associated ODE not only describes the denoising process from $\xT$ to $\xzero$ but also its reverse from $\xzero$ to $\xT$, which is achievable by solving the ODE backward.
However, the fidelity of this inversion relies on the smoothness assumption $d_\theta(\xt) \approx d_\theta(\xtone)$, which is unlikely to hold true when solving the ODEs with just a few discretization steps.
For tasks requiring multiple back-and-forth evaluations between $\xT$ and $\xzero$, an alternative inversion method is available \cite{Wallace2023-ja}.


\subsection{Motion representation}
The relative-root representation~\cite{Guo2022-vx} has been widely adopted for text-to-motion diffusion models \cite{Tevet2023-ih}. The representation is a matrix of human joint features over the motion frames with shape $D \times M$, where $D=263$ and $M$ are the representation size and the number of motion frames, respectively.
Each motion frame represents root relative rotation and velocity, root height, joint locations, velocities, rotations, and foot contact labels.
As the relative representation abstracts away the absolute root location, it can improve the generalization of motion models but makes the controllable generation more challenging \cite{Tevet2023-ih, Karunratanakul2023-nb}.

\section{\fullname}


A straightforward way to obtain a motion that fulfills a criterion $\mathcal{L}(\x)$ is by optimizing $\x^* = \argmin_{\x} \mathcal{L}(\x)$ via iterative optimization such as gradient descent. 
However, optimization in this representation space often yields implausible results as most motion samples $\x \in \mathbb{R}^{D \times M}$ do not encode plausible motions.
This motivates performing optimization on an expressive \emph{latent} space $\z$ which provides valid motion samples when decoded. 
The new optimization task becomes
\begin{equation}\label{eq:latent_opt}
    \z^* = \argmin_{\z} \mathcal{L}( f(\z) )
\end{equation}
 Such optimization has led to the success in image editing of GANs \cite{Karras2019-rv, He2019-lr},
 enabled by its smooth latent space $\z$ whose mapping is parameterized by a powerful generative model $f(\z)$ and learned from a large dataset.
 In this work, we show that the latent optimization extends well to $f$, parameterized as pretrained diffusion models in the motion domain. 

For diffusion models, our choice of the latent variable is
the diffusion's noise at time $T$, $\z = \xT$.
While $\xT$ is not perfectly smooth \cite{Preechakul2022-ql, Song2021-ac}, $\xT$ offers the highest-level abstraction readily available.
Now, getting the output $\x$ from a diffusion model $d(\cdot)$ requires solving an ordinary differential equation (ODE) using an ODE solver \cite{Song2021-yj}.
We now see Equation $\ref{eq:latent_opt}$ as
\begin{equation}\label{eq:method}
    \xT^* = \argmin_{\xT} \mathcal{L}( \mathrm{ODESolver}(d(\cdot), \xT) )
\end{equation}
This allows us to approach many motion tasks simply by adjusting the task-specific criterion $\mathcal{L}$ while keeping the motion model, $d$, intact. In this work, we use the DDIM-ODE \cite{Song2021-ac} and its solver.

This optimization is iteratively solved using gradient descent.
Starting from a certain noise $\xT$, we solve the ODE, arrive at a prediction $\x$, and evaluate the criterion function $\mathcal{L}(\x)$.
Then, we obtain the gradient $\nabla_{\xT}\mathcal{L}(\x)$ by backpropagating through the ODE solver.
An optimizer updates $\xT$ based on the gradient, possibly with a small random perturbation to encourage exploration \cite{Karras2019-rv, Wallace2023-ng}.
We repeat this until convergence.
The final output is the motion obtained by solving the ODE one last time with the optimized $\xT$. 
We call the above algorithm \fullname~(\methodname) and summarize it in Algo. \ref{algo:method}.

Maintaining the intermediate activations for solving the ODE during backpropagation can be memory-intensive. This issue can be addressed with gradient checkpointing \cite{Clark2024-bv} or an invertible ODE \cite{Wallace2023-ja, Wallace2023-ng}, at the cost of more computation or model complexity.
In the motion domain, impressive results can be achieved from an ODE solver with a minimal number of function evaluations and feasible memory overhead.
Given the ongoing efforts to reduce diffusion sampling steps, as evidenced in distillation \cite{Salimans2022-um, Song2023-vx, Liu2024-ia} and high-order solvers studies \cite{Lu2022-wv,Lu2022-rz,Zheng2023-lu}, our simple design, despite requiring back-propagation through all steps, can become increasingly relevant and applicable.

\myparagraph{Optimization through ODE solver.} Empirically, gradients via backpropagation through the ODE solver have norms spanning multiple orders of magnitude, making the optimization unstable and slow. 
We instead propose to normalize the gradient to have a unit norm, which optimizes faster in practice.
Normalized gradient methods are also found to help escape saddle-point in the loss landscape faster \cite{Murray2019-cv, Cortes2006-lp}.
When combined with momentum, the update becomes a moving average of directions, which is robust to gradient norm outliers.


\begin{algorithm}
\caption{\label{algo:method} \fullname}
\begin{algorithmic}[1]
\Require {$\xT$, motion model $d$, $\mathrm{ODESolver}$, $\mathrm{Optimizer}$, criterion $\mathcal{L}$, learning rate $\eta$, perturbation $\gamma$ (default 0)}
\While{not converged}
\State $\x \leftarrow \mathrm{ODESolver}(d(\cdot), \xT)$
\State $\nabla \leftarrow \nabla_{\xT} \mathcal{L}(\x)$ \Comment{Task-specific}
\State $\xT \leftarrow \mathrm{Optimizer}(\xT, \nabla / \| \nabla \|, \eta) + \gamma \mathcal{N}(\mathbf{0},\mathbf{1})$
\EndWhile
\State \textbf{return} $\mathrm{ODE Solver}(\xT, f)$
\end{algorithmic}
\end{algorithm}

\myparagraph{DNO fundamentally differs from a guided motion diffusion method \cite{Karunratanakul2023-nb}}
in how the criterion $\mathcal{L}(\cdot)$ is computed at each iteration. 
For guided diffusion methods, the criterion is computed on an expected $\hat{x} = \mathbb{E}[\mathbf{x}_0|\mathbf{x}_t]$ at a denoising step $t$ \cite{song2023loss,yu2023freedom} that is $\mathcal{L}(\mathbf{x}_t) = \mathbb{E}_{p(\mathbf{x}_0 | \mathbf{x}_t)} \mathcal{L}(\mathbf{x}_0) \approx \mathcal{L}(\hat{\mathbf{x}})$, resulting in severe error when $\mathrm{Var}[\mathbf{x}_0|\mathbf{x}_t]$ is large.
For DNO, the criterion is exactly computed on $\mathbf{x}$ after the full-chain denoising which eliminates the approximation error.
We further discuss these differences in the \Supp
\section{Applications}

We demonstrate the versatility of \methodname~on a wide range of conditional motion synthesis tasks. These tasks can be solved by designing a task-specific criterion $\mathcal{L}$, along with any additional auxiliary loss functions. The method for initializing the noise $\xT$ can vary depending on the task. 
In the following sections, $\x$ refers to an output from an ODE solver which is a function of $\xT$.


\subsection{Motion editing and control}
The goal of motion editing tasks is to modify a given reference motion $\x_\text{ref}$ to satisfy certain objectives. These objectives may include following a specific trajectory, conforming to specified poses in all or some keyframes, and avoiding static or moving obstacles, all while preserving the key characteristics of the input motion.

\myparagraph{Editing to follow trajectories and poses.}
Editing a motion to follow a specific trajectory or to match specific poses can be easily achieved by minimizing the average distance between each generated joint location and its corresponding target location. Target locations can be specified for any subset of joints in any subset of motion frames.

Specifically, let $\mathbf{c}^k_j \in \mathbb{R}^3$ be the target location for joint $j$ at keyframe $k$, and $\hat{\mathbf{c}}^k_j$ be its generated location from the current motion $\x$. The loss function can be defined as:
\begin{equation} \label{eq:L_gen}
    \mathcal{L}_\text{pose}(\x, O) \defeq \frac{1}{|O|} \sum_{(j,k) \in O}^{} \left\| \hat{\mathbf{c}}^k_j(\x) - \mathbf{c}^k_j \right \|_1,
\end{equation}
where $O$ is the \emph{observed} set containing $(j, k)$ pairs denoting the target joints and keyframes determined by the task.

\myparagraph{Editing to avoid obstacles.}
Obstacles can be represented as a signed distance function (SDF), whose gradient field defines the repelling direction away from the obstacles. The loss function can incorporate a safe distance threshold $\tau$, beyond which the gradient becomes zero such that
\begin{equation}
    \mathcal{L}_\text{obs}(\x) \defeq \sum_{j, k} -\mathrm{min}\left[ \mathrm{SDF}^k(\hat{\mathbf{c}}^k_j(\x)), \tau\right],
\end{equation}
where $\mathrm{SDF}^k$ is the signed distance function for obstacles in frame $k$, which may vary across frames in the case of moving obstacles.

\myparagraph{Preserving original characteristics.} When modifying, for example, a jumping motion to align with a certain trajectory, it is crucial to retain key aspects of the jump like its rhythm, height, and overall body coordination. This can be achieved using two means. First, we invert the reference motion $\x_\text{ref}$ using diffusion inversion  to obtain the corresponding noise sample $\xT_\text{ref} = \mathrm{ODESolver}^{-1}(d(\cdot), \x_\text{ref})$ and use $\xT_\text{ref}$ as the intial value for $\xT$ we are optimizing. 
Second, we penalize the distance between $\xT_\text{ref}$ and $\xT$ to ensure that it remains close to the reference motion during optimization:
\begin{equation} \label{eq:L_gen}
    \mathcal{L}_\text{cont}(\xT) \defeq \left\| \xT_\text{ref} - \xT  \right\|_2.
\end{equation}
Many motion editing tasks can be solved using a combination of these loss functions with a set of balancing weights
\begin{equation}
    \mathcal{L}(\cdot) = \mathcal{L}_\text{pose}(\x) + \lambda_\text{obs}\mathcal{L}_\text{obs}(\x, O) + \lambda_\text{cont}\mathcal{L}_\text{cont}(\xT)
\end{equation}
\subsection{Motion refinement and completion}
The tasks in this category is to reconstruct a motion from noisy and/or incomplete input.
This class of tasks includes completing a motion with missing frames or joints, seamlessly blending motions together, denoising a noisy motion, or any combination of these tasks.

\myparagraph{Motion refinement.} Given a noisy input motion, motion refinement seeks to enhance 
the input so that it becomes more realistic. 
We can solve this problem simply by starting the optimization from a random $\xT \sim \mathcal{N}(\mathbf{0}, \mathbf{I})$, setting the observed set $O$ to include all joints from the input noisy motion, and optimizing $\mathcal{L}_\text{pose}(\x, O)$. While using $\mathcal{L}_\text{pose}$ here may seem counterintuitive, as it seeks to match the predicted motion to the original noisy input, \methodname~ is able to generate a plausible motion by eliminating the noise components from the motion. 


Since this optimization begins with a random noise $\xT$, the initial prediction can be far from the desired predicted motion, requiring significant changes on $\xT$ with several optimization steps, which tend to correlate neighboring noises and reduce the representation capacity. 
We empirically find that regularizing the noise to decorrelate each latent dimension alleviates the foot skating problem. Inspired by StyleGAN2 \cite{Karras2019-rv}, we introduce a latent decorrelation loss across motion frames: 
\begin{equation} \label{eq:L_gen}
    \mathcal{L}_\text{decorr}^{m}(\xT) =
    \frac{1}{mD} \sum_{i=1}^\text{m} \mathbf{x}_{T,m}(i)^\top \mathbf{x}_{T,m}(i+1).
\end{equation}
We apply this loss at several scales of temporal resolutions $m \in \{M, M/2, M/4, \dots, 2\}$. Specifically, starting with the original length $M$, we downsample the sequence's temporal resolution by half via average pooling two consecutive frames successively.
This loss, summed over resolutions, encourages more plausible motions and can be used together with $\mathcal{L}_\text{pose}$ for motion refinement:
%
%
\begin{equation}
\mathcal{L}(\cdot) = \mathcal{L}_\text{pose}(\x, O) + \lambda_\text{decorr} \sum_{m}\mathcal{L}_\text{decorr}^{m}(\xT)
\label{motion_refinement_loss}
\end{equation}
%
%
%
%
%
%
%

\myparagraph{Motion completion.}
Unlike motion refinement, where its input contains complete joint information for all frames, this task involves taking an incomplete motion, that may be noisy, as input and seek to fill in the missing information.
We begin the optimization with $\xT \sim \mathcal{N}(\mathbf{0}, \mathbf{I})$ and apply the same loss in Equation \ref{motion_refinement_loss}, but with $O$ containing the existing joints in the input motion for $\mathcal{L}_\text{pose}(\x, O)$ term.

\section{Experiments}
\myparagraph{Datasets.}
When applicable, we evaluate generated motions on the HumanML3D \cite{guo2022t2m} dataset, which contains 44,970 motion annotations of 14,646 motion sequences from AMASS \cite{mahmood2019amass} and HumanAct12 \cite{guo2020action2motion} datasets.

\myparagraph{Motion diffusion model.}
{As \methodname~is plug-and-play and model-agnostic, it works with any trained motion diffusion model. 
When paired with MDM \cite{Tevet2023-ih}, which is trained on the HumanML3D dataset, we name the combination \methodname-MDM.
We retrained MDM with Exponential Moving Averaging \cite{Ho2020-ew} with no further modifications to help stabilize the model, which was previously found to produce inconsistent output between checkpoints.

\myparagraph{\methodname~implementation details.}
We use Adam optimizer \cite{kingma2014adam} with a learning rate of $0.05$, a linear warm-up for the first $50$ steps, and a cosine learning rate schedule to zero for the entire optimization.
We use a unit-normalized gradient. 
We set the coefficients $\lambda_\text{cont} = 0.01, \lambda_\text{decorr} = 10^3, \lambda_\text{obs} = 1.0$, and the perturbation amount $\gamma=0$.
This design choice is explored in Section~\ref{sec:ablation}.

For editing tasks, we obtain the initial $\xT$ from DDIM-100 inversion on MDM \cite{Tevet2023-ih} without text conditions. 
The subsequent optimization is run for $300$ steps, which takes approximately $3$ minutes on an Nvidia 3090 GPU.
For refinement tasks, the optimization is run for $500$ steps.
\methodname~optimization with DDIM-10 and a batch size of $16$ requires $18$~GB of GPU memory.
Optimization in all tasks is also done without text conditions to the diffusion model.
Gradient checkpointing \cite{Chen2016-vo,Clark2024-bv} can be used with DNO to reduce memory usage further for a low-memory GPU at the cost of more computation.
\subsection{Motion Editing}

\begin{table} 
\setlength\tabcolsep{3pt}
\caption{Motion editing evaluation on specific actions generated from MDM given the text prompts. We focus on actions that can be distinctly classified per frame basis. The Content Preservation scores are computed against the inputs.  
}
\vspace{-0.5em}
\footnotesize
\centering
\begin{tabular}{lcccc}
\toprule
Action  & \multicolumn{1}{p{1.2cm}}{\centering Content $\uparrow$ \\ Preserve}  & \multicolumn{1}{p{1.3cm}}{\centering Objective $\downarrow$ \\ Error (m)} & \multicolumn{1}{p{1.4cm}}{\centering Foot $\downarrow$ \\ skating ratio } & Jitter $\downarrow$\\ 
 \midrule
 ~ ``jumping" \\ 
 Input             & 1.00 & 1.62 & 0.01 & 1.42 \\
 GMD~\cite{Karunratanakul2023-nb}               & 0.64 & 0.22 & 0.12 & 3.07 \\
\methodname-MDM~Edit   & \textbf{0.95} & \textbf{0.00} & \textbf{0.05} & \textbf{1.31} \\ 
 \midrule
 ~ ``doing a long jump" \\ 
 Input             & 1.00 & 3.03 & 0.01 & 1.20 \\ 
 GMD               & 0.59 & 0.22 & 0.15 & 4.10 \\
\methodname-MDM~Edit   & \textbf{0.92} & \textbf{0.00} & \textbf{0.07} & \textbf{1.34} \\ 
 \midrule
 \multicolumn{5}{l}{~``walking with raised hand"}\\ 
 Input             & 1.00 & 2.76 & 0.01 & 0.24 \\ 
 GMD               & 0.65 & 0.16 & 0.11 & 0.51 \\
\methodname-MDM~Edit   & \textbf{0.92} & \textbf{0.00} & \textbf{0.05} & \textbf{0.32} \\ 
\midrule
 ~``crawling" \\
 Input             & 1.00 & 1.83 & 0.06 & 0.47 \\
 GMD               & 0.79 & 0.15 & \textbf{0.04} & 2.76 \\
\methodname-MDM~Edit   & \textbf{0.94} & \textbf{0.00} & 0.08 & \textbf{0.48} \\ 
\bottomrule
\end{tabular}
\label{table:result_motion_editing}
\vspace{-1em}
\end{table}





Given an input motion, we generate 16 edited motions using DNO, each with a single randomly chosen keyframe between frames 60 and 90 (i.e., 2-3 seconds after the start). The objective is to change the pelvis position to a random target location within an area ranging from -2m to 2m with respect to the initial position. We conducted this experiment on 6 input motions, resulting in a total of 96 edited motions. To generate the input motions, we use the prompt ``\emph{a person is [action]}" with a predetermined \emph{[action]}. During editing, \methodname~does not leverage the text prompt or the action class.



\myparagraph{Evaluation metrics.}
We focus on three main aspects: (1) is the output motion realistic?, (2) can it fulfill the editing objective?, and (3) how much are the original motion characteristics preserved after editing?

To evaluate the realism of the output, we measure \textit{Jitter} and \textit{Foot skating ratio}, following prior work \cite{yi2022physical, Karunratanakul2023-nb, guo2022t2m}. \textit{Jitter} is a proxy for motion smoothness, measuring the mean changes in acceleration of all joints over time in $10^2m/s^3$. \textit{Foot skating ratio} is a proxy for the incoherence between the trajectory and human motion, which measures the proportion of frames in which a foot skates for more than a certain distance (2.5 cm) while maintaining the contact with the ground (foot height $<$ 5 cm).

To evaluate the success in the editing task, we use \textit{Objective error}, which measures the distance between the target location of the edited joint at a specific frame and the corresponding joint position in the output motion. When the editing is performed on the ground plane, this metric measures the 2D distance.
%


While metrics leveraging action classifiers can provide a coarse evaluation of whether the action class is preserved or not after the editing, we require a more precise approach to quantify the content similarity. We report the ratio of frames in which both motions, before editing and after, perform the same action throughout the entire sequence, which we call \textit{Content preservation ratio}. The criteria to determine if two actions are the same are action-dependent.
The ratio will be $1$ if both motions perform the same action in all frames. 
Since there are no established methods for per-frame content similarity, in this work, we consider only distinct actions for which the boundaries of the actions can be clearly defined:
For ``jumping'', the action is defined as having both feet more than 5cm above the ground. 
For ``raised hands'', both hands need to be above the head.
For the ``crawling" motion, the head needs to be below 150 cm.

\myparagraph{Results.} As shown in Table \ref{table:result_motion_editing}, our proposed \methodname~outperforms GMD \cite{Karunratanakul2023-nb}, the state of the art in spatially-conditioned motion generation, in all actions and metrics, except for \textit{Foot skating ratio} on ``crawling.'' Overall, our method achieves significantly lower \textit{Jitter}. Furthermore, our higher \textit{Content preservation ratio} suggests that our approach retains the original content if editing is not necessary while successfully achieving the editing objectives, as indicated by the low \textit{Objective error}s.

We present qualitative results of the edited motions across various tasks in the supplementary video.
To follow a desired trajectory, we can use a set of target locations for the pelvis. Similarly, our \methodname~can edit motions to avoid obstacles without requiring explicit target keyframes. Furthermore, our method enables fine-grained editing by targeting individual joints such as the hand joint or an entire pose at a specific keyframe.
Please refer to the \Supp for additional qualitative results.


\begin{figure}
\centering
\includegraphics[width=0.5\textwidth]{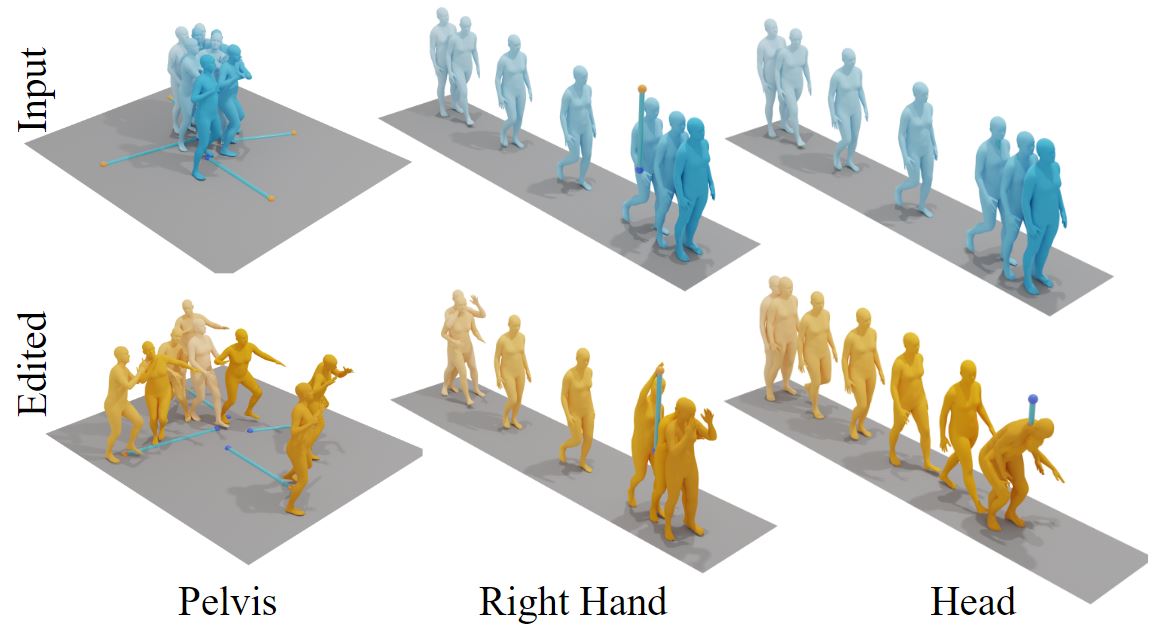} 
\caption{ Qualitative results from motion editing task. Each line indicates the starting and target location of the selected joint at a specific keyframe.}
\label{fig:result:refinement}
\vspace{-1em}
\end{figure}


\begin{table} 
\caption{Noisy motion refinement results (noise std. 5 cm) on a subset of HumanML3D \cite{guo2022t2m} dataset. 
All experiments were run with $N=300$. FIDs are computed against \textit{Real} except \textit{Real}'s FIDs which are computed against a holdout set from the dataset. HuMoR* means we exclude the sequence when its optimization fails.
DNO-MLD* runs with 1,000 optimization steps.
}
\vspace{-0.5em}
\footnotesize
\centering
\begin{tabular}{lccccc}
\toprule
 & \multicolumn{1}{p{1.7cm}}{\centering MPJPE $\downarrow$ observed (cm) } & FID $\downarrow$ & \multicolumn{1}{p{1.5cm}}{\centering Foot $\downarrow$ \\ skating ratio } & Jitter $\downarrow$ \\ 
 \midrule
Real                    & 0.0     & 0.50      & 0.08      & 0.50 \\ 
 \midrule
 \textbf{All joints} \\
 \midrule
Noisy                   & 11.1      & 58.76      & 0.66      & 28.60 \\ 
HuMoR* \cite{Rempe2021-hp}                   & 7.2      & 0.87     & 0.13      & \textbf{0.33} \\ 
GMD                     & 25.7      & 6.91      & 0.08      & 0.81 \\ 
\methodname-MDM         & 9.1      & 0.69      & \textbf{0.07}      & 0.36 \\ 
\methodname-MLD*         & 10.41    & 0.27     & 0.11      & 1.37 \\ 
\methodname-GMD         & \textbf{7.0}     & \textbf{0.10}     & 0.08      & 0.89 \\ 
\midrule
\textbf{Six joints} \\
\midrule
Noisy                   & 11.6      & 58.73      & 0.66      & 28.61 \\ 
HuMoR*                   & 8.8      & 1.40      & 0.10      & \textbf{0.20} \\ 
GMD                     & 31.2      & 7.07      & 0.08      & 0.80 \\ 
\methodname-MDM         & 8.8      & 1.15     & 0.07      & 0.38 \\ 
\methodname-MLD*         & 11.4      & 0.56      & 0.10   & 1.35 \\ 
\methodname-GMD         & \textbf{7.5}     & \textbf{0.36}     & 0.07      & 0.97 \\ 
\midrule
\textbf{Eight joints} \\
\midrule
Noisy                   & 11.4      & 58.73      & 0.66      & 28.61 \\ 
HuMoR*                   & 8.5      & 1.19      & 0.14      & \textbf{0.20} \\ 
GMD                     & 29.9      & 6.99      & 0.08      & 0.80 \\ 
\methodname-MDM         & 9.3      & 0.85      &  \textbf{0.07}     & 0.36 \\ 
\methodname-MLD*         & 11.6      & 0.42      & 0.11      & 1.36 \\ 
\methodname-GMD         & \textbf{7.2}     & \textbf{0.13}     & 0.08      & 0.96 \\ 
\midrule
\textbf{Ten joints} \\
\midrule
Noisy                   & 11.3      & 58.73      & 0.66      & 28.61 \\ 
HuMoR*                   & 9.3       & 1.06      & 0.13      & \textbf{0.21} \\ 
GMD                     & 28.4      & 6.88      & 0.09      & 0.80 \\ 
\methodname-MDM         & 9.0      & 0.66      & 0.08      & 0.39 \\ 
\methodname-MLD*         & 11.0       & 0.43      & 0.11      & 1.35  \\ 
\methodname-GMD         & \textbf{7.1}     & \textbf{0.12}     & \textbf{0.07}      & 0.96 \\ 
\bottomrule
\end{tabular}
\label{table:result_motion_projection}
\vspace{-2em}
\end{table}

\subsection{Motion Refinement}
%

We evaluate motion refinement performance on a random subset of size 300 from the test set of the HumanML3D dataset in two setups.
First, we assess if \methodname~can recover a clean and plausible motion from a noisy input motion by adding Gaussian noise with a standard deviation of 5 cm to all axes of all joints. 
We then evaluate the motion refinement performance for an input motion with incomplete and noisy joint information in three scenarios: (1) \textit{six joints}: head + two hands + two legs + pelvis, (2) \textit{eight joints}: six joints + two shoulders, (3) \textit{ten joints}: eight joints + two knees. 

\myparagraph{Evaluation metrics.} As in the motion editing tasks, we report \emph{Foot skating ratio} and \emph{Jitter} as well as the established \emph{Mean per-joint pose error} (MPJPE) between the ground-truth and predicted motions, and \textit{FID} which measures the distance between the ground-truth and synthetic data distributions using a pretrained motion encoder \cite{guo2020action2motion}.

\myparagraph{Results.} 
As shown in Table \ref{table:result_motion_projection}, we observe that, in every experiment, \methodname~successfully improves the signal by reducing MPJPE beyond the input level and producing smooth and realistic motions.
We demonstrate that \methodname's performance scales with the base model by pairing it with three motion diffusion models (ordered by FID from high-to-low on motion generation): MDM \cite{Tevet2023-ih}, MLD \cite{chen2022mld}, and GMD \cite{Karunratanakul2023-nb}.
The \methodname-GMD combination outperforms the weaker combinations, \methodname-MLD and \methodname-MDM, and the optimization-based motion prior HuMoR \cite{rempe2021humor}. And any \methodname~combination outperforms the guided diffusion method, GMD, in the motion refinement task.
For the GMD baseline, we use the spatially-conditioned generation as in the original paper. While it also successfully denoises the motion (low Jitter $0.8$), it struggles to satisfy fine pose signals and can only follow trajectory guidance leading to high MPJPE $> 25$ cm. We also tried SDEdit \cite{Meng2022-bn} on motion refinement, however, it cannot deal with such a level of noise while being able to retain the original motion. 
Additional details and tasks such as motion completion, blending, and in-betweening, are discussed in the supplementary. Note that, unlike the task-specific methods such as \cite{Du2023-kh,jiang2022avatarposer}, our method is never trained specifically for these tasks.

\subsection{Ablation studies}\label{sec:ablation}

\begin{table} 
\setlength\tabcolsep{2.8pt}
\caption{Ablation study on the noisy motion refinement task (std. 1 cm) results on a subset of HumanML3D \cite{guo2022t2m}. 
All experiments were run with $N=300$. FIDs are computed against \textit{Real} except \textit{Real}'s FIDs which are computed against a holdout set from the dataset.
}
\vspace{-0.5em}
\footnotesize
\centering
\begin{tabular}{lccccc}
\toprule
 & \multicolumn{1}{p{1.1cm}}{\centering MPJPE $\downarrow$ \\ all (cm)} & FID $\downarrow$ & Jitter $\downarrow$ & \multicolumn{1}{p{1.4cm}}{\centering Foot $\downarrow$ \\ skating ratio }\\ 
 \midrule
 Real                   & 0.0 & 0.50   & 0.50  & 0.08     \\ 
 Noisy                  & 6.4 & 9.91   & 5.87 & 0.15     \\ 
 \midrule
\methodname                                     & 8.7 & 0.66   & 0.33  & 0.07     \\  
 $-$ Normalized grad.                           & 30.2      & 4.13   & 0.34     & 0.06     \\
 $-$ Norm grad. \& $\mathcal{L}_\text{decorr}$  & 25.5      & 4.02   & 0.34     & 0.06     \\
 $-$ $\mathcal{L}_\text{decorr}$                & 6.8      & 0.65   & 0.40     & 0.10     \\ 
 $-$ LR scheduler \& warmup                     & 8.4     & 0.58    & 0.37     & 0.07     \\ 
 \midrule
 Perturb $\gamma = 0$ (\methodname)     & 8.7 & 0.66   & 0.33  & 0.07    \\ 
 Perturb $\gamma = 2 \times 10^{-4}$    & 8.5      & 0.75   & 0.34     & 0.07     \\ 
 Perturb $\gamma = 5 \times 10^{-4}$    & 8.4      & 0.73   & 0.36     & 0.07     \\ 
 Perturb $\gamma = 10^{-3}$             & 9.1      & 0.91   & 0.34     & 0.07     \\ 
 \midrule
 Optimize for 300 steps                 & 12.1      & 1.48  & 0.31     & 0.07     \\ 
 Optimize for 500 steps (\methodname)   & 8.7       & 0.66  & 0.33  & 0.07     \\  
 Optimize for 700 steps                 & 7.2      & 0.54   & 0.36     & 0.07     \\ 
 \midrule
 DDIM 5 steps                   & 9.8      & 0.90   & 0.36     & 0.09     \\ 
 DDIM 10 steps (\methodname)    & 8.7      & 0.66   & 0.33  & 0.07     \\  
 DDIM 20 steps                  & 7.9      & 0.92   & 0.34     & 0.07     \\ 
\bottomrule
\end{tabular}
\label{table:result_albation}
\vspace{-1em}
\end{table}

To motivate and justify \methodname's design choices, we conducted experiments on a surrogate motion refinement task, where we add noise with a standard deviation of $1$ cm to an input motion and seek to recover the original motion.

\myparagraph{Normalized gradients.}
As $\mathrm{ODESolver}(\cdot)$ involves iterative calls to the diffusion model, the gradients can become highly unstable.
In our experiments, we observe that the gradient norms can span multiple orders of magnitude, leading to slow or poor convergence. Table \ref{table:result_albation} demonstrates that our choice of using normalized gradients has improved the solutions by reducing MPJPE (which corresponds to the lower $\mathcal{L}_\text{pose}$ value) from $30.2$ to $8.7$ cm.



\myparagraph{Decorrelating noise.}
\methodname's main motivation is to identify a latent sample that is capable of generating a plausible motion while fulfilling the task-driven constraints. However, in practice, not all latent samples yield a plausible motion sample. One of the factors accounting for this is the uncorrelated random noise samples that the diffusion models are trained with. Correlated latent samples often results in poor motion samples.   
For example, the latent sample with all zeros has the highest likelihood, yet often generates low-quality motions. 
Our $\mathcal{L}_\text{decorr}$, motivated by this observation, discourages correlations across the time axis in $\xT$.
Table \ref{table:result_albation} demonstrates that this loss directly contributes to the motion quality. It significantly improves the \textit{Foot skating ratio} from $0.10$ to $0.07$ and the \textit{Jitter} from $0.40$ to $0.33$, though at the cost of an increased optimization difficulty as indicated by the rise in MPJPE from $6.8$ to $8.7$.
However, while it is rather straightforward to improve the MPJPE by increasing optimization steps, it is more challenging to fix the artifacts in the generated motion.

\myparagraph{Random perturbation.}
As suggested in multiple studies \cite{Wallace2023-ng, Karras2019-rv}, random perturbation is theoretically motivated as a facilitator of exploration, which helps optimization escape from local minima. 
To ensure that $\gamma$ converges to zero, we tie $\gamma$ with the learning rate scheduler, which has a warm-up period and a cosine decay. We found that small $\gamma$'s may decrease the optimization errors only marginally, while a larger $\gamma = 10^{-3}$ has a negative impact on the optimization, yet they all produced motions with worse FIDs. 
While the results are not exhaustive, they serve as a piece of evidence to support our choice of $\gamma = 0$. 

\myparagraph{ODE solver steps.} 
We experimented with varying the number of DDIM sampling steps from 5, 10, 20, which produce MPJPE scores of $9.8$, $8.7$, and $7.9$, respectively. This suggests that higher DDIM steps may be able to capture finer motion detail.  
We chose 10-step sampling as a balanced choice between result quality and resource usage.
\section{Discussion and Limitations} \label{sec:conclusion}
In this work, we proposed \methodname~a simple and versatile method to use a pretrained motion diffusion model as universal motion priors. We show that \methodname~can leverage the motion priors to achieve precise and fine-grained control for motion editing.
Apart from editing, we demonstrate that our formulation can be extended to a wide range of motion tasks.
Our method is plug-and-play and does not require training a new model for every new task.
\methodname~is, however, not a perfect motion priors. It works better when the observation has good coverage of the human body. While the decorrelation loss does help, there are situations where \methodname~cannot easily project to a realistic motion.
Ultimately, the effectiveness of our method is limited by the performance of the underlying diffusion model; nevertheless, we expect the performance of the base model to increase as the community is searching for better model designs and collecting more training data.
As the inference speed of the diffusion model keeps increasing, we hope to overcome the speed limitation of our optimization framework to achieve interactive motion editing.
In summary, our extensive studies on diffusion noise optimization effectiveness can serve as a useful basis for leveraging existing diffusion models to solve a wider range of tasks.


\myparagraph{Acknowledgement.}
{\small
This work was supported by the SNSF project grant 200021 204840.
Konpat Preechakul is funded by DARPA MCS.
We sincerely thank Bram Wallace for an insightful discussion on EDICT and DOODL which inspired this project.}

{
    \small
    \bibliographystyle{ieeenat_fullname}
    \bibliography{main,paperpile}
}

\clearpage

\begingroup

\appendix
\twocolumn[
\begin{center}
\Large{\bf Optimizing Diffusion Noise Can Serve As Universal Motion Priors \\ **Appendix**}
\end{center}
]

\counterwithin{table}{section}
\counterwithin{figure}{section}
\setcounter{page}{1}


\newpage
\section{Motion completion}
Table \ref{table:supp_result_motion_projection} shows the results of the motion completion task. The task is evaluated under the same setting as the motion refinement task in the main paper, except that the ground truth joint locations are given without added noise. The goal of the task is to generate the full-body motion given partial joint observations.

The results are consistent with the motion denoising experiment where \methodname's performance scale with base model, \methodname-GMD outperforms other baselines regarding MPJPE and FID, while HuMoR tends to produce smoother motions.

\begin{table}[h] 
\caption{Motion completion results on a subset of HumanML3D \cite{guo2022t2m} dataset. 
All experiments were run with $N=300$ except DNO-MLD* which runs with 1,000 optimization steps. FIDs are computed against \textit{Real}. The \textit{Real}'s FIDs are computed against a holdout set from the dataset.
HuMoR* means we exclude the sequence when its optimization fails.
}
\vspace{-0.5em}
\footnotesize
\centering
\begin{tabular}{lccccc}
\toprule
 & \multicolumn{1}{p{1.7cm}}{\centering MPJPE $\downarrow$ observed (cm) } & FID $\downarrow$ & \multicolumn{1}{p{1.5cm}}{\centering Foot $\downarrow$ \\ skating ratio } & Jitter $\downarrow$ \\ 
 \midrule
Real                    & 0.0     & 0.50      & 0.08      & 0.50 \\ 
 \midrule
\textbf{Six joints} \\
\midrule
HuMoR*                  & 8.7      & 1.53      & 0.13      & \textbf{0.17} \\ 
GMD                     & 31.1      & 7.08      & 0.08      & 0.79 \\ 
\methodname-MDM         & 8.5      & 1.31      & 0.07      & 0.33 \\ 
\methodname-MLD*        & 11.0     & 0.67      & 0.10      & 1.29 \\ 
\methodname-GMD         & \textbf{6.6}     & \textbf{0.30}     & 0.07      & 0.92 \\ 
\midrule
\textbf{Eight joints} \\
\midrule
HuMoR*                  & 8.4      & 1.22      & 0.13      & \textbf{0.17} \\ 
GMD                     & 29.8      & 7.06      & 0.08      & 0.79 \\ 
\methodname-MDM         & 8.7      & 0.98      & 0.07      & 0.34 \\ 
\methodname-MLD*        & 11.3      & 0.51     & 0.11     & 1.30 \\ 
\methodname-GMD         & \textbf{6.6}     & \textbf{0.12}     & 0.08      & 0.93  \\ 
\midrule
\textbf{Ten joints} \\
\midrule
HuMoR*                  & 8.3      & 1.06      & 0.12      & \textbf{0.18} \\ 
GMD                     & 28.4      & 6.88      & 0.08      & 0.79 \\ 
\methodname-MDM         & 8.6      & 0.80      & 0.07      & 0.36 \\ 
\methodname-MLD*        & 11.3     & 0.49      & 0.11      & 1.31 \\ 
\methodname-GMD         & \textbf{6.5}     & \textbf{0.11}     & 0.07      & 0.93 \\ 
\bottomrule
\end{tabular}
\label{table:supp_result_motion_projection}
\vspace{-1em}
\end{table}

\section{Additional motion-related tasks}

Under the DNO framework, the same method presented in Algorithm 1 can be adapted to many motion-related tasks without retraining the model.
In this section, we present different settings of DNO for motion blending and motion in-betweening tasks.
The qualitative results are presented in our supplementary video.

\subsection{Motion Blending.}
For motion blending, the goal is to smoothly transition from one distinct action to another. The inputs are two motion sequences and the expected output is a long motion that combines the two input motions together.
With DNO, the problem can be formulated in the same manner as the motion refinement and completion task (Sec. 5.2), where the joint locations of the concatenated input motions are used as targets and the optimization is initialized from a random $\xT \sim \mathcal{N}(\mathbf{0}, \mathbf{I})$.
To facilitate a smooth blending between motions, we define a 10-frame window around the concatenated frame as a transition period where we drop all target joints.
Consequentially, the model needs to fill in this transition according to its motion prior.
We set the content criterion $\lambda_\text{cont} = 0.0, \lambda_\text{decorr} = 10^3$, the perturbation amount $\gamma=0$, and the optimization step to 1000 for this task.

\subsection{Motion In-betweening.}
For motion in-betweening, the inputs are the starting pose and the ending pose, given by the location of each joint. The goal is to generate the in-between motion according to those two poses.
Similar to motion blending and motion completion, this task can be formulated as an optimization with partial observation as targets.
We use the same setting as in the motion blending task with the only difference being the number of target joints.


\section{Why we do not report FID for motion editing.}
As the motion Fréchet inception distance (FID) is a measurement \textit{between two data distributions}, it requires a large number of samples in both datasets \cite{guo2020action2motion}. For the motion editing task, only one motion sequence exists before editing and only a few sequences exist after editing, thus there are not enough data points to measure a meaningful FID.

\section{GMD implementation details}
To compare with GMD \cite{Karunratanakul2023-nb}, we use the released model with Emphasis projection and Dense gradient propagation for all tasks. The trajectory model is not used. When conditioned on the ground locations, we use the provided point-to-point imputing method
until $t=20$ as suggested in their experiments. The guidance is provided using the same criterion terms used in our method for all tasks.
As GMD does not support editing while preserving the content, in the editing task, we instead provided the text prompt together with the target condition as inputs for the motion editing task.
The observed joints are used without text conditioning for noisy motion refinement and motion completion.

\section{HuMoR implementation details}
We use the officially released version of HuMoR \cite{rempe2021humor} which uses both the pose prior and motion prior for evaluations.
We note that the released model is trained on a subset of the AMASS dataset \cite{mahmood2019amass} at 30 FPS which does not entirely overlap with the 20 FPS sequences in the HumanML3D dataset \cite{guo2022t2m}. The HuMoR code accepts the FPS number and does its interpolation to match the input with its learned motion prior.
We also noticed that HuMoR optimization fails on some sequences in the test set, resulting in NaN error. We removed those sequences when computing the metrics for HuMoR.


\section{SDEdit on motion refinement}
We include SDEdit \cite{Meng2022-bn} results on the motion refinement task in Tab. \ref{table:sdedit_projection}. 
We tried all possible hyperparameters $t$ in 100 increments from 100-1000.
Except for the very extreme values of $t = 1000$, SDEdit exhibits unrealistic motions affected by the presence of noise in the original motion representation with high FID and Jitter. 
At $t = 1000$, SDEdit becomes a normal DDPM generative process, and no original content is preserved.
In all cases, SDEdit fails to preserve the original content suggested by very high MPJPE.
Note that the high FID of $29.73$ for $t=1000$ comes from the fact that the motions generated from MDM without any text prompts are heavily biased toward simple motions, e.g. standing, which do not capture the wide range of possible motions in the HumanML3D dataset. 
We conclude that SDEdit is not an effective motion refinement method.

\begin{table}[h] 
\caption{
SDEdit \cite{Meng2022-bn} results on the motion refinement task (noise std. = 5 cm.). We used the default number of repetitions $k=3$ in all of the following experiments.
}
\vspace{-0.5em}
\footnotesize
\centering
\begin{tabular}{lccccc}
\toprule
 & \multicolumn{1}{p{1.7cm}}{\centering MPJPE $\downarrow$ observed (cm) } & FID $\downarrow$ & \multicolumn{1}{p{1.5cm}}{\centering Foot $\downarrow$ \\ skating ratio } & Jitter $\downarrow$ \\ 
 \midrule
 \textbf{All joints} \\
 \midrule
Real                    & 0.0       & 0.48      & 0.08      & 0.50 \\ 
Noisy                   & 11.4      & 58.82      & 0.66      & 28.61 \\ 
SDEdit (t=100)         & 346.3      & 36.10      & 0.12      & 3.12 \\ 
SDEdit (t=200)         & 313.6      & 33.92      & 0.14      & 1.61 \\ 
SDEdit (t=300)         & 288.7      & 32.47      & 0.14      & 1.11 \\ 
SDEdit (t=400)         & 259.4      & 31.24      & 0.13      & 1.01 \\ 
SDEdit (t=500)         & 226.3      & 30.53      & 0.12      & 0.86 \\ 
SDEdit (t=600)         & 187.2      & 28.99      & 0.10      & 0.93 \\ 
SDEdit (t=700)         & 150.2      & 27.76      & 0.09      & 1.48 \\ 
SDEdit (t=800)         & 122.7      & 30.83      & 0.08      & 2.35 \\ 
SDEdit (t=900)         & 79.2      & 19.09      & 0.06      & 1.33 \\ 
SDEdit (t=1000)         & 68.2      & 29.73      & 0.00      & 0.04 \\ 
\midrule
\bottomrule
\end{tabular}
\label{table:sdedit_projection}.
\vspace{-1em}
\end{table}

\subsection{Qualitative Results}
Please check our \textbf{supplementary video} for qualitative results from DNO in all tasks including motion editing, refinement, blending, and in-betweening.

\section{Differences from guided diffusion method}
While both DNO and loss-guided or classifier-guidance diffusion methods \cite{Karunratanakul2023-nb,song2023loss,yu2023freedom,Dhariwal2021-bt} can be used to produce motion samples with specific guidance objectives, these processes are completely different.

The loss-guided or classifier-guidance diffusion method (LGD) is a \textit{sampling technique} that uses the gradient of a loss function to steer the trajectory of the diffusion sampling. 
The process is done in one full-chain sampling and outputs a \textit{sample} that follows the guidance.

In contrast, DNO is a \textit{latent optimization technique} where each optimization step involves a full-chain diffusion sampling. The output is a \textit{latent code} whose decoded sample follows the guidance.

%
The differences between DNO and LGD also have the following practical implications: 

\textbf{1) Latent optimization (DNO) does not have approximation error during guidance} because it operates on the exact output $\mathbf{x}_0$ from solving the full-chain diffusion process via an ODE solver, while in LGD, the loss criterion $\mathcal{L}(\cdot)$ is \textit{approximately} computed on an expected $\hat{x} = \mathbb{E}[\mathbf{x}_0|\mathbf{x}_t]$ as explained in Eq. 7, 8 of \cite{song2023loss} and \cite{yu2023freedom} as follows:
\begin{align*}
    \mathcal{L}(\mathbf{x}_t) &= \mathbb{E}_{p(\mathbf{x}_0 | \mathbf{x}_t)} \mathcal{L}(\mathbf{x}_0) \\
    &\approx \mathcal{L}(\hat{\mathbf{x}})
\end{align*}
The approximation error is severe when $\mathrm{Var}[\mathbf{x}_0|\mathbf{x}_t]$ is large, particularly near the beginning where $T \sim 1000$. 
This means the guidance is only effective near the end of the denoising process. 
Empirically, we observe that GMD \cite{Karunratanakul2023-nb} does not reach the targets as well compared to DNO (25.7 vs 9.1 MPJPE, Table 2).




\textbf{2) The latent space can serve as universal priors for valid motions.}
DNO can answer the question ``What is the closest valid motion to the input $x$?'' by optimizing latent $x_T$ to produce a valid motion $x_0$ that best matches the input $x$.
GMD, an LGD method, is ineffective at generating valid motions from noisy inputs as shown in the refinement task in Table 2.



\textbf{3) LGD cannot easily preserve content (Tab. 1).} As editing in LGD is equivalent to new conditional sampling with the input motion, 
it is not obvious how to specify what aspects of the input motion are to be preserved and how to preserve them with LGD.

Most recent developments in diffusion image editing operate on the latent noise space with the help of the conditional inversion process \cite{Wallace2023-ja,Huberman-Spiegelglas2023-jd}. 
This direction further bolsters the merits of DNO as a latent approach for content preservation.
The latent space naturally provides smooth transitions between valid motions; samples that are close in latent space $x_T$ are also likely to be close in motion space $x_0$.
DNO enables content-preserving editing through minimal updates on the latent space and results in a minimal change in the input motion to fulfill the objectives.


As shown in the experiments, DNO enables a wide range of tasks that require precise control, motion prior, or content preservation, which cannot be effectively solved with LGD.

\end{document}